\documentclass[journal]{IEEEtran}
\usepackage{hyperref}
\usepackage{graphicx}
\usepackage{multirow}
\usepackage{cite}
\usepackage{amsmath}
\usepackage{booktabs}
\usepackage{algorithm}
\usepackage{algorithmic}
\usepackage{color}
\ifCLASSOPTIONcompsoc
  \usepackage[caption=false,font=normalsize,labelfont=sf,textfont=sf]{subfig}
\else
  \usepackage[caption=false,font=footnotesize]{subfig}
\fi

\begin{document}

\title{STNReID: Deep Convolutional Networks with Pairwise Spatial Transformer Networks for Partial Person Re-identification}

\author{Hao Luo,
        Wei Jiang,
        Xing Fan,
        Chi Zhang,
\thanks{Hao Luo, Wei Jiang, Xing Fan are with the State Key Laboratory of Industrial Control Technology, College of Control Science and Enginneering, Zhejiang University, Hangzhou 310027, China; E-mail: haoluocsc@zju.edu.cn, jiangwei\_zju@zju.edu.cn, xfanplus@zju.edu.cn.}
\thanks{Chi Zhang is with Beijing Research Institute of Megvii Inc, Beijing, China; E-mail: zhangchi@megvii.com}
\thanks{Manuscript received May 14, 2019; revised Nov 18, 2019.}}

\markboth{Journal of \LaTeX\ Class Files,~Vol.~14, No.~8, August~2015}%
{Shell \MakeLowercase{\textit{et al.}}: Bare Demo of IEEEtran.cls for IEEE Journals}
\maketitle

\begin{abstract}
Partial person re-identification (ReID) is a challenging task because only partial information of person images is available for matching target persons.
Few studies, especially on deep learning, have focused on matching partial person images with holistic person images.
This study presents a novel deep partial ReID framework based on pairwise spatial transformer networks (STNReID), which can be trained on existing holistic person datasets.
STNReID includes a spatial transformer network (STN) module and a ReID module.
The STN module samples an affined image (a semantically corresponding patch) from the holistic image to match the partial image.
The ReID module extracts the features of the holistic, partial, and affined images.
Competition (or confrontation) is observed between the STN module and the ReID module, and two-stage training is applied to acquire a strong STNReID for partial ReID.
Experimental results show that our STNReID obtains 66.7\% and 54.6\% rank-1 accuracies on Partial-ReID and Partial-iLIDS datasets, respectively.
These values are at par with those obtained with state-of-the-art methods.
\end{abstract}

\begin{IEEEkeywords}
Partial Person ReID, STN, Occlusion, Deep learning.
\end{IEEEkeywords}

\IEEEpeerreviewmaketitle

\section{Introduction}
Person re-identification (ReID) is widely applied in video surveillance and criminal investigation applications \cite{wang2019incremental}.
It has achieved huge improvements especially on holistic person datasets using deep learning techniques in recent years \cite{Sun_2018_ECCV,luo2019alignedreid++,su2017pose,wei2018glad}.
However, partial observations of person images exist due to occlusions, viewpoints, and special tasks in real-world applications.
For example, a target person inside a subway station, airport, or supermarket may be occluded by ticket gates, baggage and checkout counters, or other things/people.
The patch of a person's body is constantly maintained by using several techniques, such as human detection or skeleton model, to reduce such information interference.
In such situations, most ReID models trained on holistic person datasets are unstable and frequently fail in searching for accurate images.
Hence, Zheng \emph{et al.} \cite{zheng2015partial} addressed the partial person re-identification (partial ReID) task and proposed the Partial-ReID dataset.
In recent years, the partial ReID has gradually attracted the attention of both researchers and engineers due to its huge application value.

\begin{figure}[tb]
\centering
\includegraphics[width=.9\linewidth]{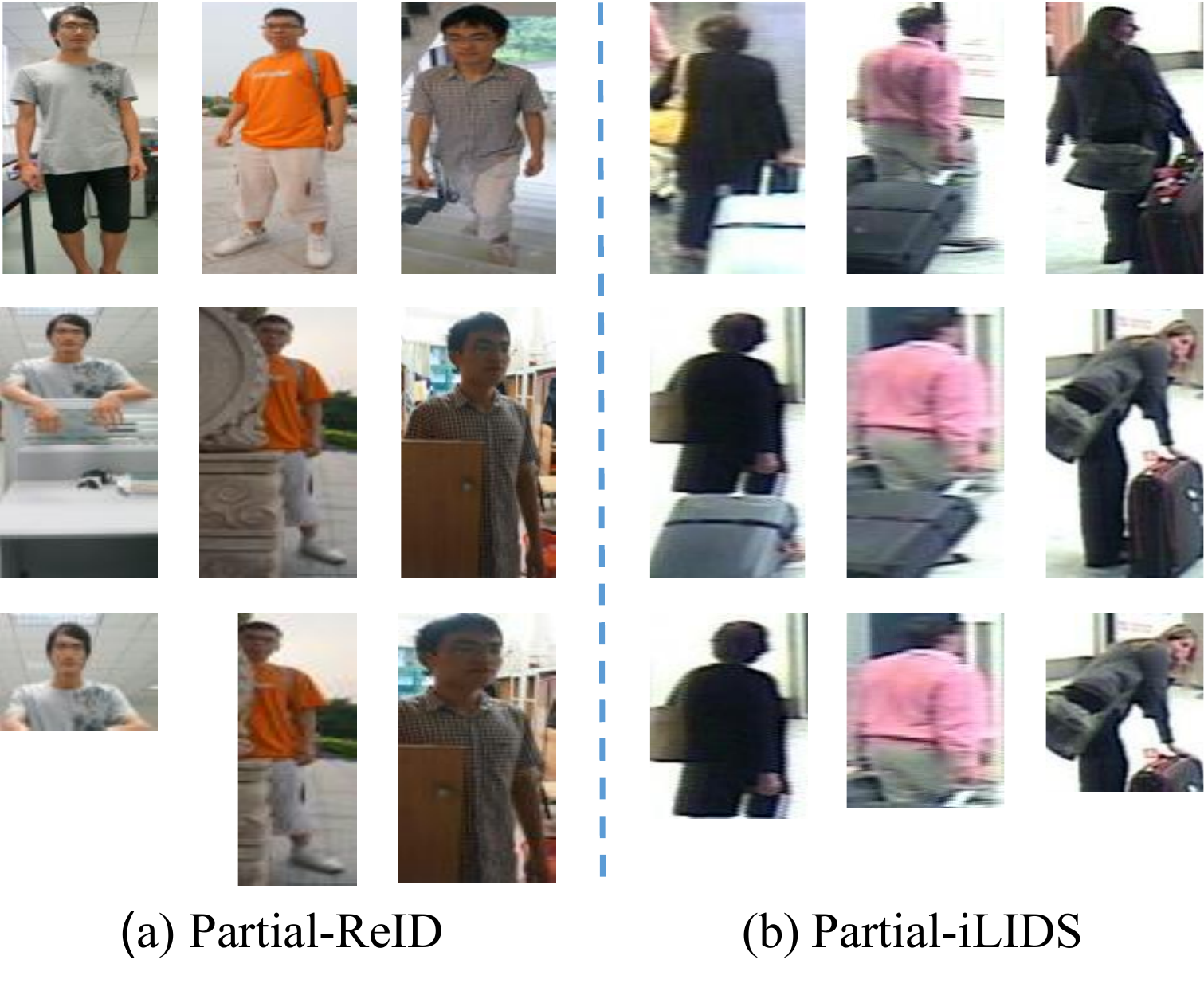}
\caption{Example of holistic person images (first row), annotated partial person images for recognition (third row), and the corresponding non-partial images (second row).}
\label{fig:demo}
\end{figure}

Partial ReID is a challenging task because only partial information is available for matching target persons (as shown in Fig. \ref{fig:demo}).
It is also made difficult by the need to determine whether a partial image should be resized to a fix-sized image.
Undesired deformation exists when the model is resized to a fix-sized image.
By contrast, the model should match an arbitrary image patch to a differently sized holistic image when the partial image is not resized.
Two features or feature maps with different sizes cannot be compared, especially on end-to-end convolutional networks (CNNs).
In \cite{he2018deep}, He \emph{et al.} summarized the drawbacks of several kinds of solutions.
For example, Sliding Window Matching (SWM) \cite{zheng2015partial} establishes a sliding window with the same size as the partial image and utilizes it to search for the most similar region on each holistic image.
However, the computational cost of this solution is expensive due to the traversal calculation.
Deep Spatial Reconstruction (DSR) \cite{he2018deep} directly matches two arbitrary-sized feature maps by sparse reconstruction in a deep learning framework to accelerate the matching process.
Although DSR is an impressive work for partial ReID, its arbitrary-sized matching mechanism limits the utilization of the tensor computing power of the GPU device.
DSR requires one-to-one matching after the extraction of the feature maps of partial and holistic images.
It also becomes inefficient with the enlargement of image size.
In CNNs, a batch is composed of several tensors (images) with the same sizes.
Specifically, CNNs should deal with ``resizing" and ``matching" problems at the same time.

In the other hard, Part-based Convolutional Baseline (PCB) \cite{Sun_2018_ECCV}, which divides the image into several horizontal stripes and concatenates multiple local features, achieves great performance in holistic person ReID task.
Some PCB based methods have been proposed for partial ReID task.
Two example methods are Spatial-Channel Parallelism Network (SCPNet) \cite{fan2018scpnet} and Visibility-aware Part-level Model (VPM) \cite{sun2019perceive}.
Such methods can solve horizontal occlusions but are hard to deal with vertical occlusions.
In addition, the alignment between partial and holistic images is rough in these methods.
Iodice \emph{et al.} proposed Partial Matching Net (PMN) \cite{iodice2018partial} that align the body parts of two images with the help of human body joints.
However, PMN needs an extra pose estimation model to detect human body joints.
Local features and semantic information can improve the performance but always need more computation.

In this study, we present a novel deep ReID framework (STNReID) based on spatial transformer networks \cite{jaderberg2015spatial} for partial ReID.
STNReID includes one spatial transformer network (STN) module and one ReID module.
The STN module utilizes high-level ReID features of partial and holistic images and predicts the parameters of 2D affine transformers, such as resizing (scaling), rotation, and reflection, of two images.
Then, the STN module samples an affined image from the holistic image to match the partial image.
The ReID module extracts the global features of the affined and partial images to retrieve the target person images.
Trained partial images are generated from holistic person datasets.
Hence, STNReID can be trained by standard holistic ReID datasets without the need for additional labeled data.
However, the performance of the STN module decreases with the increase of the power of the ReID module.
Therefore, a weak ReID is used to train a strong STN module, and the STN module is restricted to fine-tune the ReID module.
Such two-stage training enables the acquisition of a strong STNReID model.
In the inference stage, STNReID can match a certain partial image with several holistic images in a batch.
This end-to-end one-to-many matching is more efficient than one-to-one matching is.
The major contributions of our work are summarized as follows:
\begin{itemize}
\item A novel partial ReID framework (STNReID) based on STNs is proposed.
The proposed STNReID can be trained on holistic person datasets without the need for additional labeled partial images.
To our best knowledge, this pairwise STNs is involved into person ReID task at the first time.
\item Experimental results show that the STNs can be trained by the high-level semantic features and its performance is affected by the ReID module.
A two-stage training process is proposed to enable the acquisition of a strong STNReID for partial ReID.
\item The experimental results show that our STNReID achieves competitive outcomes on Partial-ReID \cite{zheng2015partial} and Partial-iLIDs \cite{zheng2011person} benchmarks.
\end{itemize}

\section{Related Works}
In this section, deep learning-based person ReID methods are summarized, and the existing relevant studies on partial ReID are reviewed because partial ReID is a sub-topic of person ReID.
Then, STNs \cite{jaderberg2015spatial} and their application to person ReID are investigated.

\subsection{Deep person ReID}
Deep learning-based person ReID uses deep CNNs (DCNNs) to represent the features of person images and surpasses most of traditional methods \cite{zheng2013reidentification,liao2015person,wang2016zero}.
On the basis of the loss functions of training DCNNs, most existing studies focus on two methods, namely, robust representation learning and deep metric learning.
Representation learning-based methods \cite{zheng2016person,Zheng_2017_CVPR} aim to learn robust features for person ReID by using the Softmax loss (ID loss).
An ID embedding network (IDENet) \cite{zheng2016person,Zheng_2017_CVPR} regards each person ID as a category of a given classification problem.
In addition, Fan \emph{et at.} \cite{fan2018spherereid} obtained the variants of the SoftMax function and achieved superior performance in the field of ReID.

Compared with representation learning, deep metric learning-based algorithms directly learn the distance of an image pair in the feature embedding space \cite{shi2016embedding}.
The typical metric learning method is the triplet loss \cite{liu2017end}, which pulls the distance of a positive pair and pushes the distance of a negative pair.
However, triplet loss is easily influenced by the selected samples.
Hard mining techniques \cite{hermans2017defense,xiao2017margin,ristani2018features} widely used to obtain triplet loss with high accuracies.
Improved triplet loss \cite{cheng2016person} and quadruplet loss \cite{chen2017beyond} are variants of the original triplet loss.
At present, the combination of ID loss with triplet loss has attracted considerable attention due to its remarkable performance.

On the other hand, researching on global and local features of person images is another important branch of person ReID.
To our best knowledge, Luo \emph{et at.} \cite{luo2019bag,Luo2019TMM} achieves the best performance with global features by bag of effective tricks.
However, global features ignore the spatial or local information in an image and encountered a performance bottleneck.
In recent years, more and more works \cite{Sarfraz_2018_CVPR,su2017pose,Sun_2018_ECCV,varior2016siamese,wei2018glad,luo2019alignedreid++,zhangintegration,zhao2017spindle,zheng2017pose,zhou2018large} have introduced into the local features.
Some works \cite{Sarfraz_2018_CVPR,su2017pose,wei2018glad,zhao2017spindle,zheng2017pose} focus on the pose or skeleton information of a person, which are applied into aligning the local features of different body parts.
These methods need extra labeled data to train an available pose estimation model.
Without extra supervision and having high efficiency, stripe or grid based methods \cite{Sun_2018_ECCV,luo2019alignedreid++,zhangintegration} are another mainstream in the field of person ReID.
In these works, local features of each stripe or grid can contain more spatial and local information and boost the performance, especially when merging with global features.
All above techniques are successful on holistic person ReID datasets such as Market1501 \cite{zheng2015scalable}, DukeMTMCReID \cite{ristani2016MTMC}, CUHK03 \cite{li2014deepreid}, MSMT17 \cite{wei2018cvpr}, \emph{etc}.

\subsection{Partial ReID}
With the development of person ReID, partial ReID has gradually attracted the attention of researchers due to its huge value in real-world ReID applications.
However, few studies have solved the task, and the performance of partial ReID is unsuitable for practical applications.
As a solution to this problem, many methods \cite{donahue2014decaf,girshick2014rich} have been designed such that they are able to directly resize an arbitrary patch of a person image to a fixed-size image and extract the fixed-length global features for matching.
However, the various scale deformations caused by such rough methods are difficult to address.
In this regard, part-based models provide an optional solution.
In \cite{zheng2015partial}, Zheng \emph{et al.} proposed a global-to-local matching model called SWM that can capture the spatial layout information of local patches and introduced a local patch-level matching method called the Ambiguity-sensitive Matching Classifier (AMC) that is based on a sparse representation classification formulation with an explicit patch ambiguity model.
However, the computation cost of the AMC+SWM is extremely expensive because the extraction of features requires considerable time without sharing computation.
Apart from handcrafted features or models, He \emph{et al.} \cite{he2018deep} used a fully convolutional network to generate spatial feature maps with certain sizes and whose pixel-level features are consistent.
Then, DSR is conducted to match a pair of person images with different sizes.
Proposing pyramid pooling layer to extract multi-scale spatial features to alleviate the influence of scale mismatching, He \emph{et al.} updated DSR into Spatial Feature Reconstruction (SFR) \cite{he2018recognizing}.
Taking advantage of pyramid features, SFR achieves better performance than DSR.
For holistic person ReID, Part-based Convolutional Baseline (PCB) \cite{Sun_2018_ECCV} concatenates several horizontal striped features to represent the full image information and has achieved great success.
Inspired by PCB, two stripe based methods, \emph{i.e.}, Spatial-Channel Parallelism Network (SCPNet) \cite{fan2018scpnet} and Visibility-aware Part-level Model (VPM) \cite{sun2019perceive}, were proposed to align a partial image to a holistic image using horizontal striped features.
Such methods can deal with horizontal occlusions but are hard to handle vertical occlusions.

\begin{figure*}[htb]
\centering
\includegraphics[width=.9\linewidth]{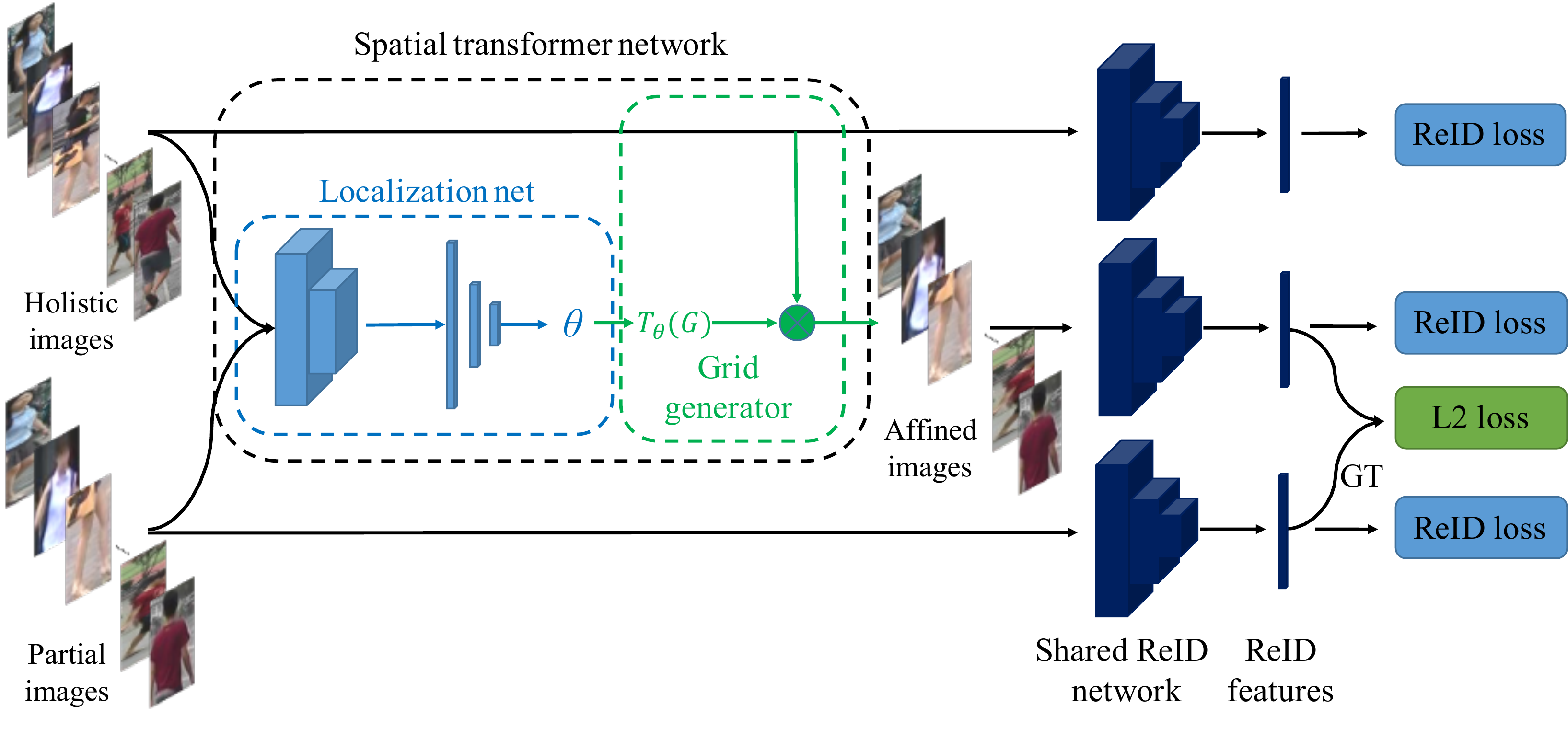}
\caption{The framework of STNReID, which includes an STN and a ReID module. The ReID loss combines ID loss and Triplet loss.}
\label{fig:Structure}
\end{figure*}

\subsection{Spatial Transformer Networks}
Spatial Transformer Networks (STNs),  which include a localization network and a grid generator, make up the deep learning method proposed in \cite{jaderberg2015spatial}.
The localization network utilizes feature maps and outputs the parameters of 2D affine transformations through several hidden layers.
Then, such predicted transformation parameters are transferred in the grid generator to create a sampling grid, which is a set of points where the input feature map is sampled to produce the transformed output.
An STN can perform 2D affine transformations, such as reflection, rotation, scaling, and translation.
For person ReID, Zheng \emph{et al.} \cite{zheng2018pedestrian} propose Pedestrian Alignment Network (PAN), which combines the STN and a deep ReID network.
However, PAN is similar to STNs as it predicts parameters on the basis of the feature maps of one image and aligns the weak spatial changes of holistic person images.
Inspired by STN, PAN, and DSR, we propose a novel partial ReID model called STNReID.
Unlike the abovementioned methods, STNReID estimates the parameters of affine transformations by analyzing a pair of holistic and partial person images.
\section{Methods}
This section introduces the architecture of the proposed STNReID and demonstrates the training of an improved STNReID in a two-stage pipeline.

\subsection{STNReID}
As shown as Fig. \ref{fig:Structure}, STNReID includes an STN module and a ReID module.
The STN module predicts the parameters of the affine transformations, and then crops patches (affined images) from holistic person images to match partial images.
The ReID module extracts the global features of holistic images, partial images and affined images.

\subsubsection{\textbf{STN Module}}

The holistic and partial images are denoted as $I_h$ and $I_p$, respectively.
In the training stage, the partial image is randomly cropped from the holistic image.
$I_h$ and $I_p$ are the tensors with the shape of $H \times W \times C$, where $H$ and $W$ the height and width of the resized images with fixed sizes, respectively; and $C$ is the number of image channel (e.g. $C=3$ for RGB images).
Then, $I_h$ and $I_p$ are concatenated as a new tensor $I_{h,p}$($H \times W \times 2C$) and is fed in the localization network of the STN module.
The localization network outputs $\theta$ through a series of hidden layers, and the parameters of transformation $T(\theta)$: $\theta = f_{loc}(I_{h,p})$.
The localization network function $f_{loc}()$ is dependent on the structure of the localization network.
\begin{equation}
\left(
 \begin{matrix}
   x_i^s  \\ y_i^s
  \end{matrix}
  \right)
  = T_{\theta}(G) = A_{\theta}
   \left(
 \begin{matrix}
   x_i^t  \\
   y_i^t  \\
   1
  \end{matrix}
  \right) =     \left[
 \begin{matrix}
   \theta_{11} &   \theta_{12} & \theta_{13} \\
    \theta_{21} &   \theta_{22} & \theta_{23}
  \end{matrix}
  \right]
    \left(
 \begin{matrix}
   x_i^t  \\ y_i^t  \\ 1
  \end{matrix}
  \right)
\end{equation}

where $(x_i^t, y_i^t)^T$ are the target coordinates of the regular grid in the affined image, $(x_i^s, y_i^s)^T$ are the source coordinates in the holistic image that define the sample points, and $A_{\theta}$ is the affine transformation matrix dependent on $\theta$.
$T_{\theta}(G)$ is the grid generator, which determines the sampling of the affined image from the holistic image.
Additional details, such as back propagation and sampling mechanism, can be found in \cite{jaderberg2015spatial}.
Affined image $I_a$ can be obtained in STN module $f_{STN}()$ as follow:
\begin{equation}
	I_a = f_{STN}(I_{h,p})
\end{equation}

In this study, the localization network, the architecture of which is presented in Table \ref{STN}, includes two convolutional layers (BN and ReLU layers), two pooling layers, and four fully connected layers (ReLU layers apart from the last layer).
Downsampling is conducted four times by using two convolutional layers. This process is performed on the basis of two considerations: (1) the spatial information is more important than the semantic information is for the STN module, and excessive convolutional layers may lose considerable spatial information; (2) the STN module requires large receptive fields to associate the relevant patches of two images.
Thus, downsampling is frequently applied. The fully connected layers estimate the 6D parameters of $\theta$ to sample the affined image from the holistic image. Such affined image is used in the ReID module.

\begin{table}[htp]\small
\caption{ Architecture of our localization network. In this study, input  $I_{h,p}$ is a tensor with $256 \times 128 \times 6$ shape.
The designed network has few layers with a large receptive field. The FC4 layer is a regression layer and is not followed by a ReLU layer.}
\begin{center}
\begin{tabular}{cccc}
\hline
Name & Output Size & Parameters & Padding  \\
\hline
\hline
Conv1 & $128 \times 64 \times 16$ & $[7\times 7, 16]$, stride=2 & (1,1) \\
Max Pool &$64 \times 32 \times 16$ & $2\times 2$, stride=2 & (0,0) \\
Conv2 & $32 \times 16 \times 32$ & $[3\times 3, 32]$, stride=2 & (1,1) \\
Max Pool &$16 \times 8 \times 32$ & $2\times 2$, stride=2 & (0,0) \\
\hline
Flatten & 4096 & - & - \\
FC1 & 512 & $4096 \times 512$ & - \\
FC2 & 128 & $512 \times 128$ & - \\
FC3 & 32 & $128 \times 32$ & - \\
FC4 & 6 & $32 \times 6$ & - \\
\hline
\end{tabular}
\end{center}
\label{STN}
\end{table}

\subsubsection{\textbf{ReID Module}}

The ReID module extracts the global features of three kinds of images, namely, holistic, partial, affined images.
For convenience, $f_{ReID}()$ is used to denote the ReID module.
The ReID module outputs the global features and predicted person IDs in the training stage and outputs the global features only in the inference stage.
In this study, ResNet50 is used as ReID module.
In particular, for an arbitrary image $I$, we have
\begin{equation}
	(p_I, f_I) = f_{ReID}(I)
\end{equation}
where $p_I$ is the predicted logits and $f_I$ is the global feature of the image $I$.
$p_I$ and $f_I$ are used to calculate the ReID loss:
\begin{equation}
	L_{R}(I) = L_{R}(p_I, f_I) = L_{ID}(p_I) + L_{Tri}(f_I)
\end{equation}
$L_{R}$ can be any ReID loss function such as $L_{ID}$ \cite{Zheng_2017_CVPR} and adaptive weighted triplet loss $L_{Tri}$ \cite{ristani2018features}.
In this study, $L_{R}$ will be introduced in next section.

The ReID module has another important usage because it guides the STN module in reconstructing partial images.
The ReID feature of affined image should be similar to the one of the partial image in the feature space.
To achieve such target, $L_{STN}$ is expressed as follow:
\begin{equation}\label{eq:stn}
	L_{STN}(I_p, I_a) = ||f_{I_p}- f_{I_a}||_2^2
\end{equation}
where $f_{I_p}$ and $f_{I_a}$ denote the global features of image $I_p$ and $I_a$ respectively.
The loss of STNReID includes three ReID losses and $L_{STN}$, which is expressed as:
\begin{equation}\label{loss}
	L = L_{R}(I_h)+L_{R}(I_p)+L_{R}(I_a) + L_{STN}(I_p, I_a)
\end{equation}

\noindent
\textbf{Discussion:}
Although the $\theta$ is easily acquired for training data, we do not directly regress it using the grand truth $\theta$ or original image information.
Because in the inference stage or a real-world system, the partial and holistic images are come from two different images.
It is impossible to find a $\theta$ to sample a patch, which is almost same with the partial image like training pairs, from the holistic image.
However, the global feature of such sampled patch may be similar with the feature of the partial image, even there exist large pose variants.

\subsubsection{\textbf{ReID Loss}}
In this study, ReID loss $L_{R}$ is computed as $L_{ID}$ \cite{Zheng_2017_CVPR} and adaptive weighted triplet loss $L_{Tri}$ \cite{ristani2018features}, which is similar with \cite{luo2019bag}.
In particularly, the ReID module is feed into an image $I$ and outputs the predicted logits of person ID (denoted as $p_I$) and the global feature (denoted as $f_I$).
$p_I = \{p(1),p(2),p(3),...,p(N)\}$ is a N-dimensional vector where $N$ is the total number of person IDs in the training dataset.
$p(k)$ is the probability that the picture belongs to the $k$-th person ID.
Similarly, $q_I = \{q(1),q(2),q(3),...,q(N) \}$ is a ont-hot vector that represent the label of image $I$.
If the grand-truth ID of image $I$ is $k$, $q(k)$ will equal to $1$ and other elements will be 0.
Then, the ID loss is computed as cross entropy loss and written as:
\begin{equation}\label{IDLoss}
    L_{ID}(p_I) = -\sum_{k=1}^{N} q(k)\log{p(k)}
\end{equation}

Another loss is adaptive weighted triplet loss $L_{Tri}$ \cite{ristani2018features}.
Each batch mini-batch contains K sample images for each of P identities, \emph{i.e.} the batch size is $B = P \times K$.
For an anchor sample $x_a = I$, positive samples $x_p \in P(a)$ and negative samples $x_n \in N(a)$, the triplet loss is denoted as:
\begin{equation}\label{Tri}
    L_{Tri}(I) =\left[m+\sum_{x_{p} \in P(a)} w_{p} d(x_{a}, x_{p})-\sum_{x_{n} \in N(a)} w_{n} d\left(x_{a}, x_{n}\right)\right]_{+}
\end{equation}
where $m$ is the given inter-person separation margin, $d$ denotes distance of global features, and $[\cdot]_{+}=\max (0, \cdot)$.
$w_{p}$ and $w_{n}$ are the adaptive weights for positive and negative samples respectively:
\begin{equation}\label{Tri}
    w_{p}=\frac{e^{d\left(x_{a}, x_{p}\right)}}{\sum_{x \in P(a)} e^{d\left(x_{a}, x\right)},}, w_{n}=\frac{e^{-d\left(x_{a}, x_{n}\right)}}{\sum_{x \in N(a)} e^{-d\left(x_{a}, x\right)}}
\end{equation}
More details can be found in \cite{ristani2018features}.

\subsection{Two-Stage Training}\label{TST}
STNReID can be trained end to end in a single stage. Nevertheless, the performance of the STN module is influenced by the ReID module.
An improved model with two-stage training is demonstrated in this section.

As shown as Eq.\ref{eq:stn}, $f_{I_p}$ is used to guide the STN module's training.
However, such approach results in the poor performance of the STN module because the ReID module is powerful.
$I_p$ and $I_a$ belong to the same person ID.
$f_{I_p}$ and $f_{I_a}$ are similar in the feature space when the ReID module has a strong performance, because the features have strong clustering characteristics, especially on metric learning.
In this case, the STN module cannot easily obtain knowledge.
Table \ref{Table2} and Figure \ref{fig2} illustrate the conclusion.
Additional explanations are provided in section \ref{AFS}.

\begin{figure}[tb]
\centering
\includegraphics[width=.99\linewidth]{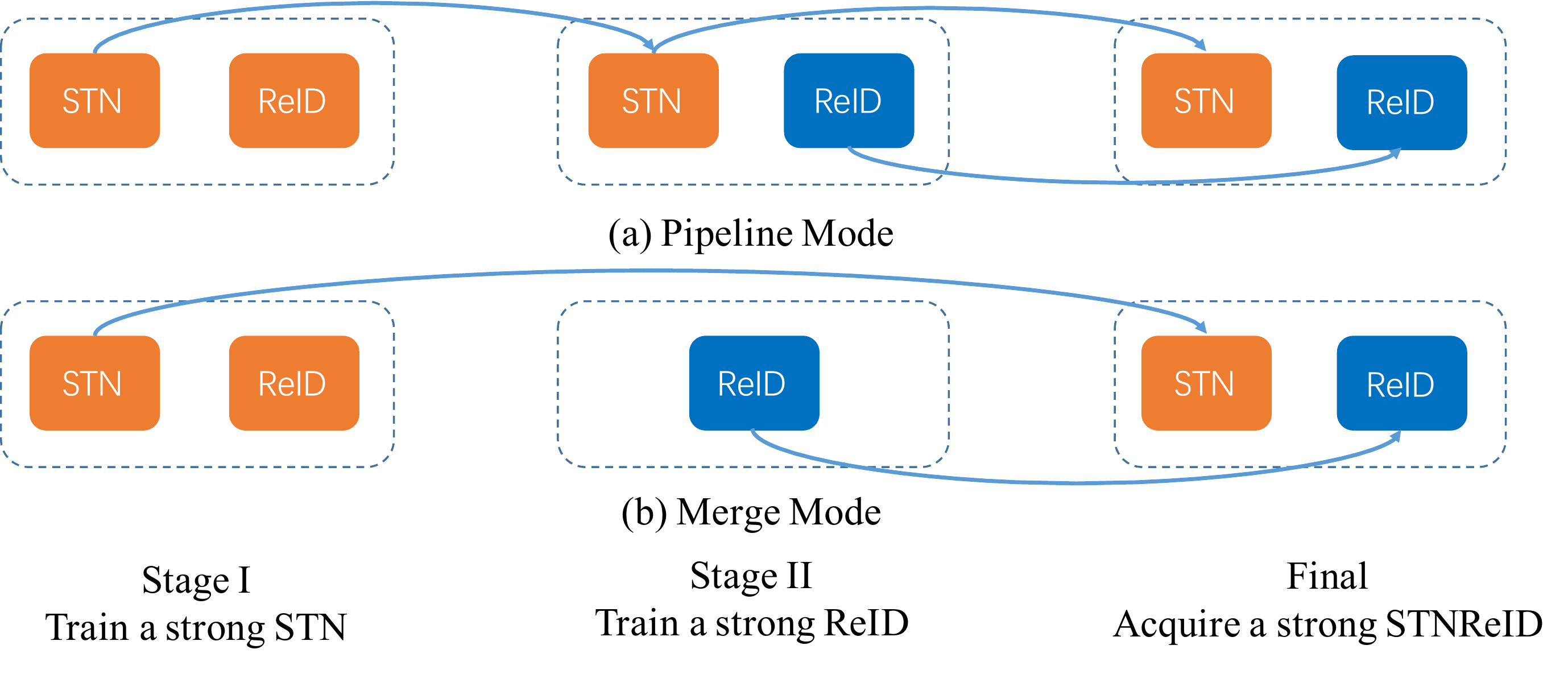}
\caption{Two-stage training mechanism of STNReID.}
\label{fig:two_stage}
\end{figure}

As shown as Figure \ref{fig:two_stage}, a two-stage training mechanism is used to solve the abovementioned problem.
In the first stage, STNReID is trained with a weak ReID module to acquire a strong STN module.
In specifically, as shown as EP1 in Table \ref{Table2}, the ReID module is trained with only ID loss, and ImageNet pre-trained weights are not used to initialize it.
In the second stage, the STN module of STNReID acquired from the first stage is be frozen.
In addition, as shown as EP5 in Table \ref{Table2}, ID loss, triplet loss and some training tricks are involved to acquire a strong ReID module.

For the second stage, there are two alternative modes called Pipeline Mode (PM) and Merge Mode (MM).
For the Pipeline mode (see Figure \ref{fig:two_stage}(a)), the ReID module in fine-tuned with the frozen STN module.
Because of fine-tuning, STNReID(PM) is more suitable for the partial ReID task.
However, someone may have trained a very strong ReID model on large-scale holistic or partial ReID datasets.
In this case, the Merge Mode (see Figure \ref{fig:two_stage}(b)), which merges a STN module and a ReID module into STNReID, is free to expand a trained ReID model to an STNReID(MM) model.

The commonality between PM and MM is training a strong STN module at first, and then acquiring a strong ReID module.
Finally, two-stage training mechanism let us get a better STNReID for the partial ReID task.

\section{Experimental Settings}

\noindent
\textbf{Datasets.} One holistic person ReID dataset and two partial person ReID datasets are introduced in this study.
Market1501\cite{zheng2015scalable} is the most popular holistic dataset, and includes 32668 holistic person images of 1501 person IDs.
The training set consists of 12,936 images of 751 identities.
Only the training set is used to train the model in our experiments.
Then trained models are evaluated on two partial ReID benchmarks: Partial-ReID \cite{zheng2015partial} and Partial-iLIDs \cite{zheng2011person}.
Partial-ReID includes 600 images of 60 persons, with 5 full-body images and 5 partial images per person.
The images were collected at an university campus with different viewpoints, background and different types of severe occlusions (see Fig. \ref{fig:demo}).
Partial-iLIDs is a simulated partial dataset based on iLIDs \cite{zheng2011person}.
In the Partial-iLIDs dataset, there are 119 persons with 238 person images captured by multiple non-overlapping cameras.
All the images are test data for these two partial ReID datasets.
The partial and holistic images are regarded as the probe and gallery set respectively for both datasets .

\noindent
\textbf{Evaluation Protocol.} We follow the evaluation protocol in \cite{he2018deep, zheng2015partial}.
In specific, the test set is random splited into query and gallery subsets and is evaluted by normal Cumulative Match Characteristic (CMC) ten times.
Then, We average all CMC accuracies for verification experiments to evaluate our method.
Following the previous works, the models are trained on Market1501 and evaluated on partial ReID datasets.

\noindent
\textbf{Generate Partial Images.} To train a STNReID, we should generate partials images from holistic images.
Firstly, we randomly choose a direction to crop the holistic image.
Then, we randomly crop $20\% \sim 60\%$ information of holistic images in the selected direction to simulate the partial images.
These generated partial images have two usages, as the input partial images to train the STNReID and as augmentation to train strong baselines for partial ReID.

\noindent
\textbf{Baseline.}
We boost the baseline applied in \cite{he2018deep}, because some algorithms may fail on a strong baseline.
Firstly, we use the adaptive weighted triplet loss proposed in \cite{ristani2018features}, which can improve approximately 3\% rank-1 accuracy compared with triplet loss with hard example mining.
Secondly, label smooth trick \cite{szegedy2016rethinking} is a practice to reduce overfitting for softmax (ID) loss, which can also improve more than 2\% rank-1 accuracy.
Finally, we use generated partial images as training images for the data augmentation, which is important for partial ReID methods.
Such augmentation can provide the model robust deformation adaptability, and some methods mainly focus on solving the deformation problem.
In above ways, our baseline can achieves 58.2\% rank-1 accuracy on Partial-ReID with only being trained on Market1501.
Since DSR only obtained 39.3\% rank-1 accuracy and did not report the performance of its baseline, our baseline surpass the baseline of DSR by a large margin.
In addition, our baseline achieves 93.1\% rank-1 accuracy on Market1501 with standard evaluation protocol.
In overall, it is a strong baseline for person ReID task.

\noindent
\textbf{Settings.}
Our backbone network is ResNet50 implemented by the open source\footnote[1]{\href{https://github.com/L1aoXingyu/reid_baseline}{https://github.com/L1aoXingyu/reid\_baseline}}.
Each image is resized into $256\times128$ pixels.
Data augmentation involves random horizontal flipping, cropping, and generated partial images. The margin of triplet loss is 0.3, and the batch size is set to 32 with 4 randomly selected images for every 8 identities.
The Adam optimizer is used, and the learning rate is $2\times10^{-4}$ for the first 150 epochs and decays to $2\times10^{-5}$ for next 150 epochs.
Label smooth method \cite{szegedy2016rethinking} is applied to avoid overfitting.
The weight reduction is set to $5\times10^{-4}$.
Parameters $\theta$ are initially set to $[1,0,0,0,1,0]$.

\section{Experimental Results}
\subsection{The ReID module affects the STN module}\label{AFS}
When end-to-end traning the STNReID, the confrontation between the STN module and the ReID module is presented, as previously discussed in Section \ref{TST}.
In this section, several experiments are conducted to verify such assumption.

\begin{figure}[tb]
\centering
\includegraphics[width=.9\linewidth]{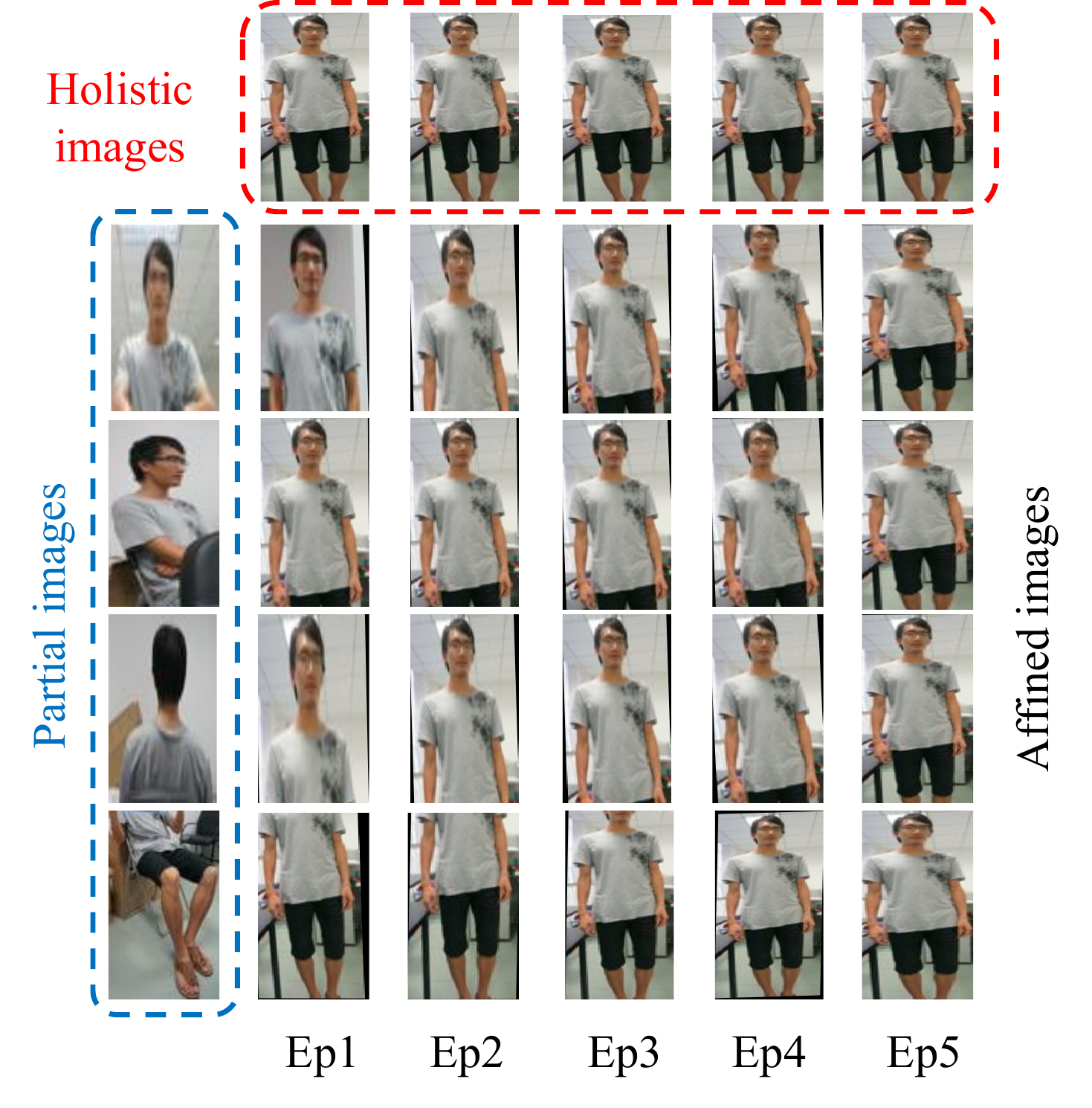}
\caption{Sample affined images of five experiments. A holistic image and four different partial images are used. The affined images of each column belong to one certain experiment. The affined images of each row show the comparison between different experiments. }
\label{fig2}
\end{figure}

To acquire ReID modules with different performances, we conduct five experiments to train the STNReID models. The training settings are shown in Table \ref{Table2}. However, the performance of the STN module is difficult to independently evaluate. Several aspects are considered. In particular, the rank-1 accuracy of STNReID models is assessed on partial ReID datasets. Then, the STN modules are removed and re-evaluated, that is, only the performance of ReID modules is assessed. A performance gap is observed between such two test settings.
The performance of the ReID modules from Ep1 to Ep5 strengthens gradually, and the improvements from the STN modules are smaller. However, the rank-1 accuracies should not be considered strictly because they are a normal condition in which the improvement by the algorithm becomes small given a strong baseline model. Therefore, several affine images generated by the five STN modules are visualized in Figure \ref{fig2}. The STN module of Ep1 accurately matches four partial images and the holistic image. The third matching pair of Ep2 slightly exhibits some redundant information. In addition, the STN module of Ep5 cannot learn much knowledge, and the affined images of the last column are similar to the holistic image.

\begin{figure*}[htb]
\centering
\includegraphics[width=.99\linewidth]{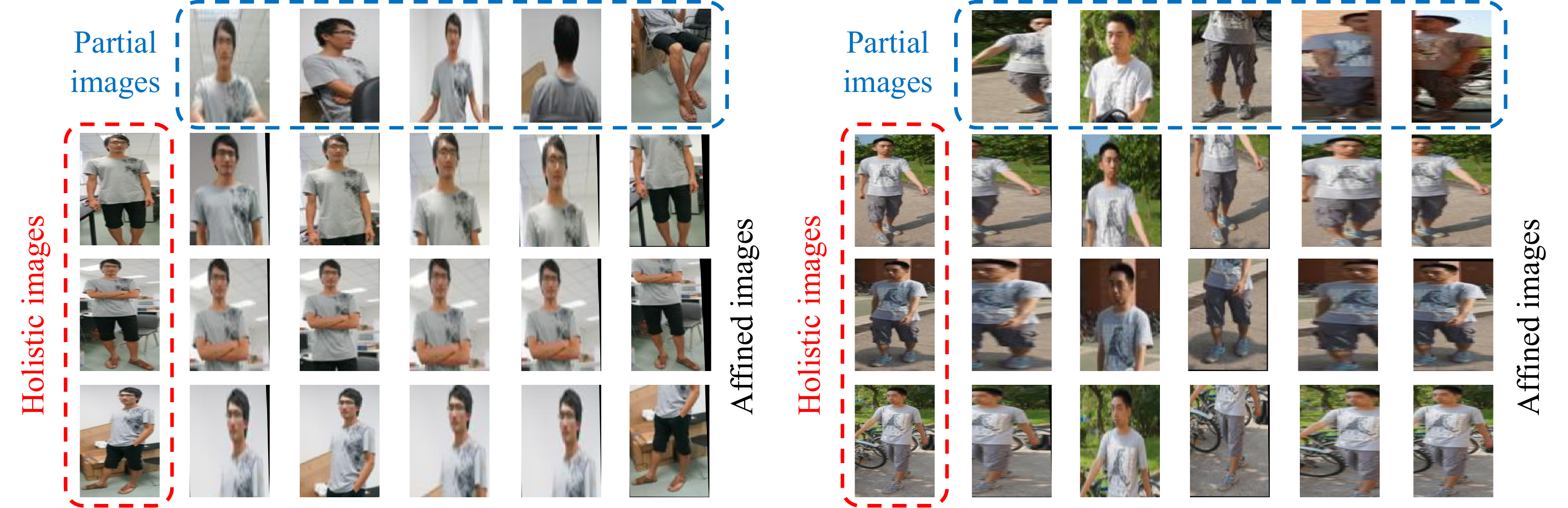}
\caption{The example affined images of the strong STN module. We choose three holistic images and five partial images of two identities.}
\label{fig3}
\end{figure*}

\begin{figure}[htb]
	\centering
	\includegraphics[width=.99\linewidth]{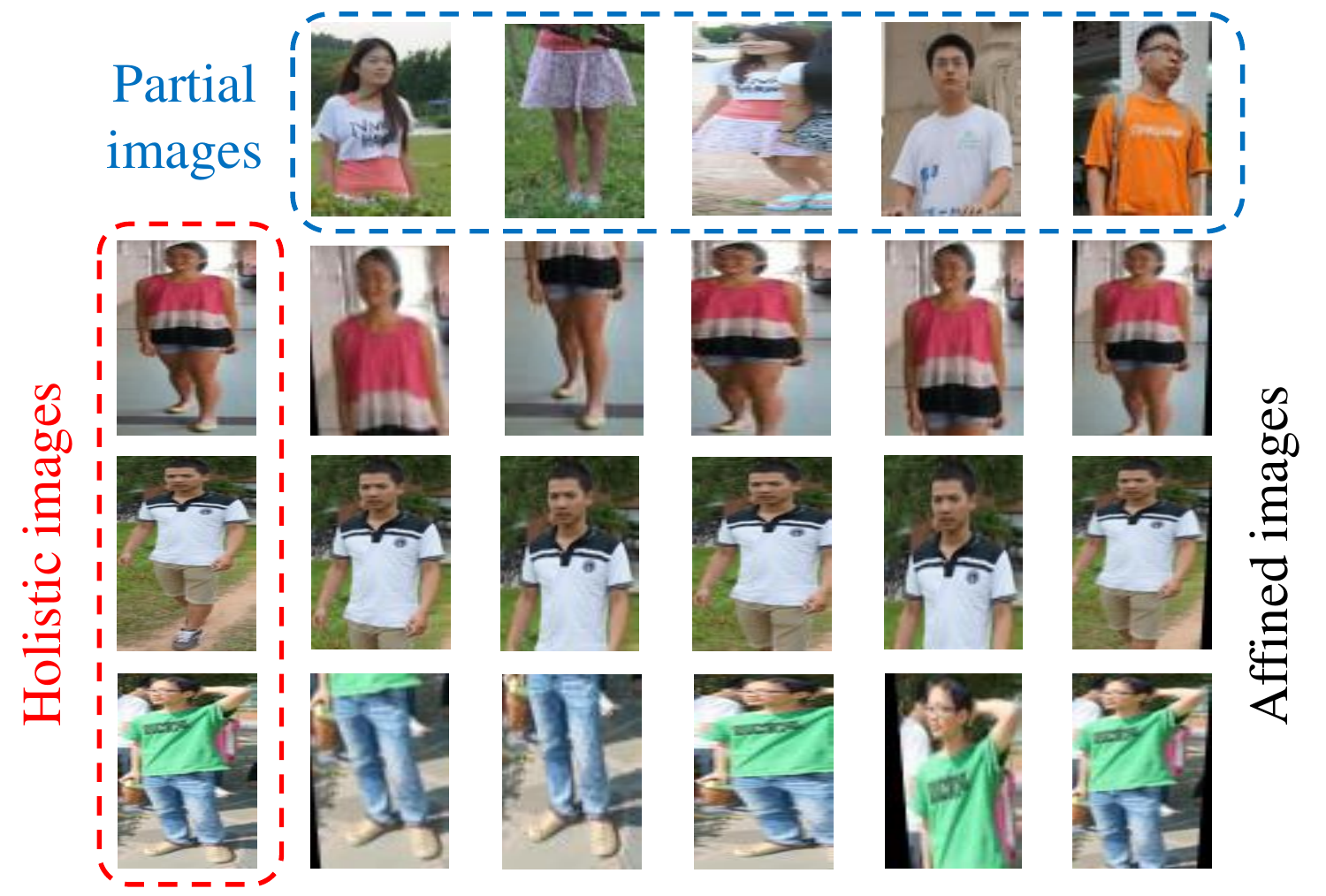}
	\caption{The example affined images of negative pairs between different IDs.}
	\label{neg}
\end{figure}

In conclusion, confrontation is observed between the STN module and the ReID modules. A strong ReID model is unsuitable for training a strong STN module in an end-to-end STNReID model with only one stage. 
We can acquire a strong STN module through the second stage training with a weak ReID model. 
Some example affined images generated by such strong STN module are show in Figure \ref{fig3} and \ref{neg}.
In Figure \ref{fig3}, both the partial image and holistic image belong to the same IDs.
Whether vertically or horizontally partial images, we can get roughly accurate results. The model sometimes refused by `left' and `right', because the direction information is difficult to distinguish.
We futher show some examples between different IDs in Figure \ref{neg}.
The STN module did not see any negative pairs in the training stage, so it mainly sample regions according global features and body-part information from holistic person images.
Although the affined images are sometimes not accurate, the ranking performance is not influenced a lot, because the model cannot find a matching region between a negative pair.

\begin{table}[tp]\footnotesize
\caption{Training settings of five experiments. `PT' indicates the use of pre-trained weights on ImageNet to initialize the ReID module. `LS' denotes the label smooth trick. `w/o STN' stands for the performance of the ReID module without the STN module. `Imp' denotes the improvement of rank-1 accuracy from the STN module on Partial-ReID.}
\begin{center}
\begin{tabular}{c|c|c|c|c|c|c|c}
\hline
 & PT& LS & $L_{ID}$ & $L_{Tri}$  & w/o STN & STNReID & Imp\\
\hline
Ep1 &$\times$& $\times$ & $\surd$ & $\times$ & 34.4 &41.7 & +7.3\\
Ep2 &$\surd$& $\times$ & $\surd$ & $\times$ & 39.5&46.3 & +6.8\\
Ep3 &$\surd$& $\surd$ & $\surd$ & $\times$ & 49.3& 53.3 & +4.0\\
Ep4 &$\surd$& $\times$ & $\surd$ & $\surd$ & 59.5& 60.5 & +1.0\\
Ep5 &$\surd$& $\surd$ & $\surd$ & $\surd$ & 62.9&63.6 & +0.8\\
\hline
\end{tabular}
\end{center}
\label{Table2}
\end{table}

\subsection{Ablation studies of two-stage training}

\begin{figure}[h]
	\centering
	\includegraphics[width=.48\linewidth]{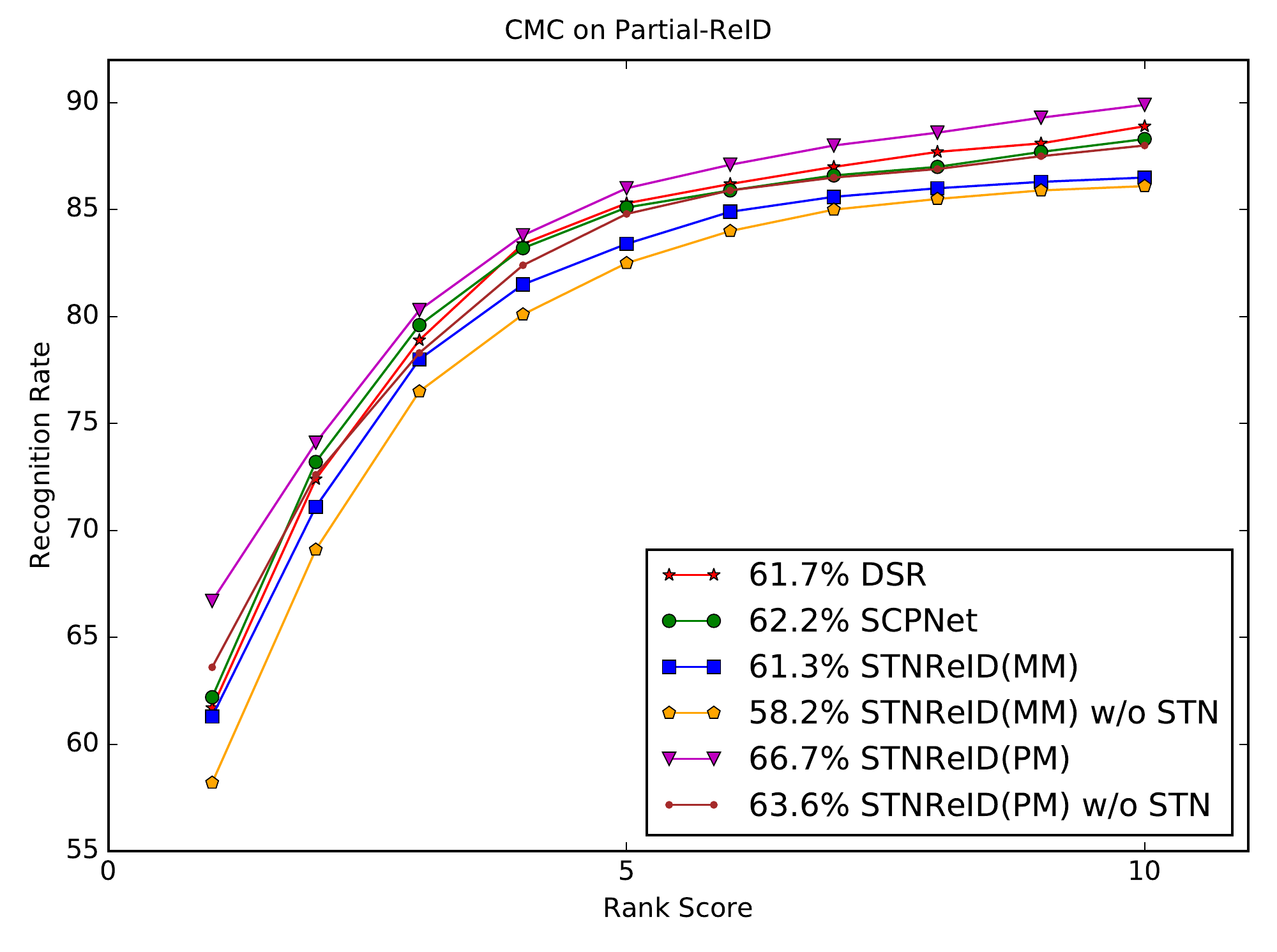}
	\includegraphics[width=.48\linewidth]{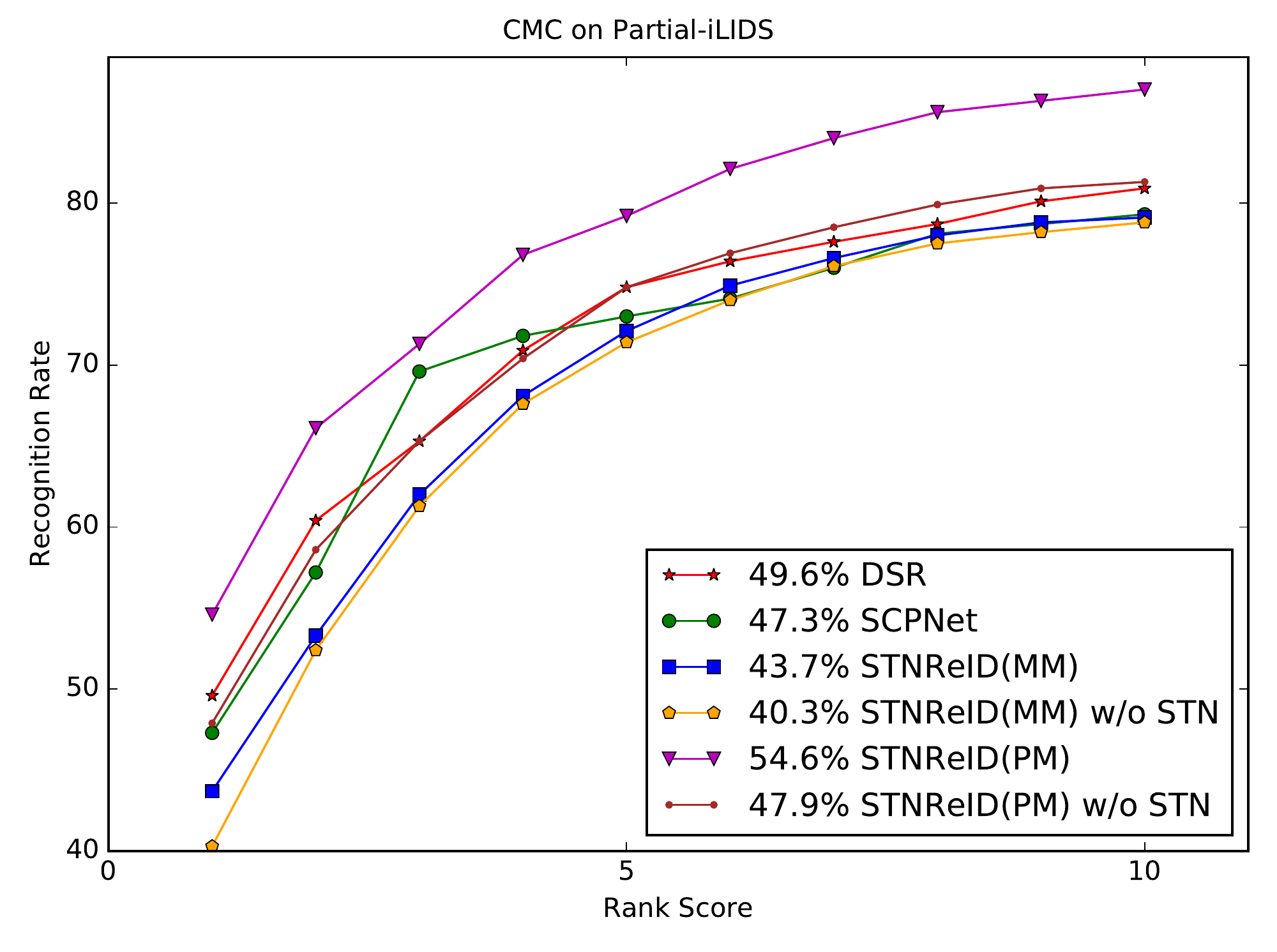}
	\caption{CMC curves on Partial\_ReID and Partial\_iLIDS.}
	\label{cmc}
\end{figure}

In this section, STNReID in the second stage is analyzed, along with the pipeline mode (PM) and merge mode (MM).
The STN module obtained from Ep1 is initialized and frozen in the second stage to train STNReID models.
The results are presented in Table \ref{ablation} and Figure \ref{cmc}.
The ReID backbone (Baseline) obtains 58.2\% and 40.3\% rank-1 accuracies on the Partial-ReID and Partial-iLIDS datasets, respectively.

For the MM, STNReID(MM) directly merges the ReID backbone and the frozen STN module.
This mode can improve rank-1 accuracies by 3.1\% and 3.4\% on Partial-ReID and Partial-iLIDS datasets, respectively.
For the PM, STNReID (PM) achieves 66.7\% and 54.6\% rank-1 accuracies on the Partial-ReID and Partial-iLIDs datasets, respectively; thus, it outperforms the baseline and STNReID (MM) by large margins.
With the removal of the STN module, the rank-1 accuracies of non-affine features are reduced by 3.1\% and 6.7\% on the Partial-ReID and Partial-iLIDS, respectively.
STNReID (PM) performs better than STNReID (MM) does on the partial ReID task because it fine-tunes the ReID module on partial images.
Nevertheless, STNReID (MM) can be used in certain situations, such as when applications have a strong ReID model trained on a large-scale dataset, which requires considerable time and computing resources.
STN module training is inexpensive.

\renewcommand{\multirowsetup}{\centering}
\begin{table}[t]\small
  \begin{center}
  \begin{tabular}{ c|cc|cc}
\hline
    				& \multicolumn{2}{c|}{Partial-ReID} & \multicolumn{2}{c}{Partial-iLIDS}	 \\
  Model			& R-1 	& R-5	& R-1 	& R-5 	 \\
 	\hline
	\hline
 STNReID(MM) w/o STN		&58.2	&82.5	&40.3	&71.4			\\
 STNReID(MM) 	&\color{blue}{61.3}	&\color{blue}{83.1}	&\color{blue}{43.7}	&\color{blue}{72.1} \\
 STNReID(PM) w/o STN &63.6	&84.8	&47.9	&74.8			\\
STNReID(PM)		&\color{red}{66.7}	&\color{red}{86.0}	&\color{red}{54.6}	&\color{red}{78.2}			\\
\hline

  \end{tabular}
  \end{center}
  \caption{\label{ablation}Ablation studies of two-stage training. MM and PM denote merge mode and pipeline mode respectively, and w/o STN indicates the removal of the STN module in the testing stage.}
\end{table}

\subsection{Comparison to the State-of-the-Art}

The comparison of STNReID models and state-of-the-art methods is shown in Table \ref{SOTA}. Few studies have explored partial ReID. SWM and AMC are methods based on handcrafted features.
DSR is the first deep learning-based method submitted in CVPR2018. Therefore, the comparison of our models and DSR is highlighted here.
The reported rank-1 accuracies of DSR are 39.3\% and 51.1\% for the Partial-ReID and Partial-iLIDS datasets, respectively.
Our STNReID (PM) obtains 66.7\% and 54.6\% rank-1 accuracies for the Partial-ReID and Partial-iLIDS, respectively.
Because the training of our baseline is different from the training of DSR's baseline.
DSR is evaluated with our baseline for fair comparison.
The average image size of Partial-iLIDS is $363.3 \times 138.3$, and is thus twice or thrice larger than that of Partial-ReID.
So that DSR method performs better on Partial-iLIDS.
The performance of DSR is better than that of STNReID (MM) but is worse than that of STNReID (PM).

SCPNet is similar with STNReID which represents the image with size $256 \times 128$ by a global feature.
However, SCPNet takes advantage of spatial striped features and reports the performance of the model trained on four holistic ReID datasets.
We used its open-sourced code to train the model on Market1501.
Finally, SCPNet achieves 42.6\% and	37.0\% rank-1 accuracies on Partial-ReID and Partial-iLIDS respectively.
For comparison, SCPNet is also evaluated with our baseline and obtains 62.2\% and 47.3\% rank-1 accuracies for the Partial-ReID and Partial-iLIDS, respectively.

SFR and VPM both are both are pre-print papers on ArXiv and are feed into the image with size $384 \time 192$.
Our STNReID evaluates the models with images of size $256 \times 128$.
So SFR and VPM may achieve great performance with local features on Partial-iLIDS because of larger image size.
As shown as Table \ref{Time}, the image size of Partial-ReID is smaller than $256 \times 128$ while the ones of Partial-iLIDS is larger than $256 \times 128$.
Because Resizing a small image into a larger size will not increase information, but resizing a large image into a smaller size will result in lossing information.
SFR and VPM do not obtain a great improvement on Partial-ReID.
In addition, our STNReID uses global features while SFR and VPM use multi local features.
Image size is more sensitive to local features than global features.
Enlarging image size dose not provide a significant improvement for STNReID.
It is note that SFR involve in multi-scale local features and VPM apply an attention mechanism apart from aligning partial person image to a holistic ones.
Both they are effective for person ReID task with more computation.
Finally, since SFR and VPM are excellent methods and achieve better performance on Partial-iLIDS, our STNReID is a competitive method for partial ReID task.

\renewcommand{\multirowsetup}{\centering}
\begin{table}[tb]
	\begin{center}
		\begin{tabular}{ c|ccc|ccc}
			\hline
			& \multicolumn{3}{c|}{Partial-ReID} & \multicolumn{3}{c}{Partial-iLIDS}	 \\
			Methods				& R-1 	& R-3	& R-5	& R-1 	& R-3	& R-5 	 \\
			\hline
			\hline
			
			Resizing model			&19.3	&32.7 & 40.0	&21.9	&37.0	& 43.7		\\
			SWM                     &24.3	&45.0 & 52.3	&33.6	&47.1	& 53.8		\\
			AMC                     &33.3	&46.0 & 52.0	&46.9	&64.8	& 69.6		\\
			AMC+SWM(ICCV15)         &36.0	&51.0 & 60.0	&49.6	&63.3	& 72.3		\\
			DSR(CVPR18)$^*$\cite{he2018deep}		    &39.3	&55.7 &65.7	&51.1	&61.7  & 70.7\\
			SCPNet(ACCV18)\cite{fan2018scpnet}          &42.6	&56.9 &66.1	&37.0	&45.7  & 58.0\\
			SFR(Arxiv18)$^+$\cite{he2018recognizing}	    &56.9	&78.5 &-	&63.9	&74.8  & -\\
			VPM(ArXiv19)$^+$\cite{sun2019perceive}           &67.7	&81.9 &-	&65.5	&74.8  & -\\
			STNReID(PM)$^+$		&66.4	&81.0 &86.2	&56.1 &73.3	&81.5	\\
			\hline
			Baseline				&58.2	&76.5 	&82.5	&40.3 	&61.3	&71.4			\\
			DSR(Our)			    &61.7	&78.9	&85.3	&49.6 	&65.3	&74.8			\\
			SCPNet(Our)            &62.2	&79.6     &85.1	&47.3	&63.5  & 73.0\\
			STNReID(MM) 		    &61.3	&76.8     &83.1		&43.7	&62.6 	&72.1 \\
			STNReID(PM)			&\color{red}{66.7}	&\color{red}{80.3} &\color{red}{86.0}	&\color{red}{54.6} &\color{red}{71.3}	&\color{red}{79.2}	\\
			\hline
		\end{tabular}
	\end{center}
	\caption{\label{SOTA}Comparison to state-of-the-art methods on Partial-ReID and Partial-iLIDS. $^*$ means the reported results in original paper. However, our baseline is different from DSR's baseline and SCPNet's baseline. So DSR and SCPNet are evaluated on our baseline for fair comparison, which is denoted as DSR(Our) and SCPNet(Our) respectively. $^+$ means the evaluated models are feed into images with size $384 \times 192$. They achieve great performance on Partial-iLIDS because of larger image size.}
\end{table}

\subsection{Extended Comparison on Holistic Perons ReID}
Since a good partial ReID framework should not only be useful on partial ReID task, but also for holistic ReID task, we compare our STNReID with some holistic ReID methods on Market1501 in Table \ref{Holi}.
Because holstic person ReID dose not need matching match a partial person image to a holistic one, the STN module will not learn any knowledge.
Finally, the STNReID achieve 93.8\% rank-1 accuracy and 84.9\% mAP, which is close to the performance of the Baseline on Market1501.
We select some popular and state-of-the-art methods, including PCB, AlignedReiD++, OSNet and ABD-Net, etc, for holistic person ReID in Table \ref{Holi}.
Our STNReID outperfroms some of them, but is not able to outperfrom the newest studies such as Bot, OSNet and ABD-Net.
However, we claim that our method is mainly proposed for partial person ReID; and we believe that our STNReID can achieve better performance with multi-scale features or attention mechanism, etc.

\begin{table}[t]\small
	\begin{center}
		\begin{tabular}{c|c|c|c}
			\hline
			Method		& Reference	& R-1& mAP	 \\
			\hline
			\hline
			DSR	\cite{he2018deep}		& CVPR18	& 83.6	& 64.3	 \\
			PCB	\cite{Sun_2018_ECCV}		& ECCV18	& 92.4	& 77.3	 \\
			PCB+RPP	\cite{Sun_2018_ECCV}		& ECCV18	& 93.8	& 81.6	 \\
			AlignedReID++	\cite{luo2019alignedreid++}		& PR19\textsf{}	& 91.9	& 76.8	 \\
			VPM	\cite{sun2019perceive}		& CVPR19	& 93.0	& 80.8	 \\
			BoT\cite{luo2019bag}		& CVPRW19	& 94.5	& 85.9	 \\
			OSNet	\cite{zhou2019osnet}		& ICCV19	& 94.8	& 84.9	 \\
			ABD-Net\cite{chen2019abd}		& ICCV19	& 95.6	& 88.3	 \\
			\hline
			STNReID		& /			& 93.8	& 84.9	 \\
			\hline
		\end{tabular}
	\end{center}
	\caption{\label{Holi} Comparison to some methods on Market1501.}
\end{table}

\begin{table}[tb]\small
	\begin{center}
		\begin{tabular}{c|c|c}
			\hline
			& Partial-ReID& Partial-iLIDS	 \\
			\hline
			Gallery Size		& $p=60$& $p=119$	 \\
			Image Size ($H \times W$)	& $151.4 \times 52.1$& $363.3 \times 138.3$	 \\
			\hline
			\hline
			DSR (1-1 Match)		& \color{blue}{0.112s}	& 3.042s	 \\
			STNReID 	(1-1 Match)	& 0.847s	& 1.899s	 \\
			STNReID 	(1-2 Match)	& 0.452s	& 1.034s	 \\
			STNReID 	(1-16 Match)	& 0.169s	& 0.336s	 \\
			STNReID 	(1-32 Match)	& 0.157s	& 0.311s	 \\
			STNReID 	(1-48 Match)	& 0.150s	& \color{blue}{0.303s}	 \\
			\hline
		\end{tabular}
	\end{center}
	\caption{\label{Time} Comparison of average time durations in person image identification (including the time cost of feature extraction). $p$ is the number of person images in the gallery set. 1-N denotes the matching of one partial image with $N$ holistic images in a batch at each time.}
\end{table}

\subsection{Computation Efficiency}

DSR and STNReID are evaluated in PyTorch on a PC with 16 GB RAM and TitanX GPU device. The average time of feature extraction and image retrieval for identifying a probe image from p gallery images is reported. For fair comparison, all computations are performed by the GPU device. For DSR, the computation costs of feature extraction and feature sparse reconstruction remarkably increase with the increase in image size. The average image size of Partial-iLIDs is $363.3 \times 138.3$ and is thus twice or thrice larger than that of Partial-ReID. As shown in Table \ref{Time}, DSR consumes 0.112s and 3.042s to identify a person image for Partial-ReID and Partial-iLIDs, respectively. However, the time cost mainly depends on gallery size $p$ due to the resizing of images to fixed sizes for STNReID. DSR can only perform one-to-one matching, which is inefficient. For STNReID, the probe image can be evaluated repeatedly for $N$ times and can be matched with $N$ different gallery images in a batch at each time. The time consumption gradually reduces with the increase of batch size $N$. For example, the time cost decreases from 1.899s ($N=1$) to 0.303s ($N = 48$) for Partial-iLIDs. Overall, DSR is effective for small images, whereas STNReID is mainly affected by gallery size. In addition, experimental results show 1-1 matching is inefficient.

\section{Conclusion, Disadvantages and Future Works}
In this paper we introduced a novel deep convolutional networks with Spatial Transformer Networks (STNReID) for partial ReID.
STNReID can sample the most similar patch from the holistic image to match the partial image.
The experimental results demonstrate the effectiveness of our method on Partial-ReID and Partial-iLIDS.
The computation costs of STNReID and DSR are also presented in Section Appendix.
Experimental results show that STN can learn knowledge from high-level semantic features.
It has the potential of solving the partial ReID task.

However, we also thought about the disadvantages of the proposed framework.
Firstly, STNReID only considers how to match the positive pairs through minimizing the L2 distances of the global features.
In the inference stage, it is hard to find the accurate association between negative sample pairs.
In addition, STNReID is a siamese network essentially.
The siamese network is not as efficient as the one-stream ReID model.
We will study how to train a stronger STNReID with a simple, direct and efficient way in the future.

\section*{Acknowledgment}
This research was funded by the National Natural Science Foundation of China under Grant 61633019, the Science Foundation of Chinese Aerospace Industry under Grant JCKY2018204B053 and the Autonomous Research Project of the State Key Laboratory of Industrial Control Technology, China (Grant No. ICT1917).

\ifCLASSOPTIONcaptionsoff
  \newpage
\fi

\bibliography{egbib}{}
\bibliographystyle{IEEEtran}
%
%
%
%
%

\end{document}